\documentclass[runningheads]{llncs}
\usepackage[T1]{fontenc}
\usepackage{graphicx}
\usepackage[utf8]{inputenc}
\usepackage[english]{babel}
\usepackage{enumitem}
\usepackage{amsmath} 
\usepackage{amssymb}
\usepackage{mwe} 
\usepackage{multirow}
\usepackage{algorithm}
\usepackage{algpseudocode}
\usepackage{comment}
\usepackage{graphicx}
\usepackage{xcolor}
\usepackage[colorinlistoftodos]{todonotes}
\usepackage{booktabs}
\usepackage{ulem}

\begin{document}
%
\title{Flow Matching Meets 3D Curvilinear Structure Segmentation in Medical Imaging}
\titlerunning{Flow Matching Meets 3D Curvilinear Structure Segmentation ...}
%
\author{Sidi Mohamed Sid'El Moctar\orcidID{0009-0009-5490-1916}, Nicolas Vitry \and Hélène Bouvrais \orcidID{0000-0003-1128-1322}} 
%
\authorrunning{SM Sid'El Moctar, N Vitry, H Bouvrais}
%
\institute{CNRS, Univ. Rennes, Institute of Genetics and Development of Rennes (IGDR), Rennes, France}
\maketitle              
\begin{abstract}
Segmentation of curvilinear anatomical structures in 3D medical images remains challenging due to complex topology, severe class imbalance, weak contrast, and large variations in structure morphology. While deep learning approaches for 3D curvilinear segmentation have been proposed, they are often tailored to specific anatomies or modalities, limiting generalization across clinical settings and leaving room for improvement. Recent generative models have shown the benefits of iterative prediction for structured segmentation tasks, yet diffusion-based methods suffer from computationally expensive sampling, hindering their use on high-resolution 3D volumes.
We present 3D-CurvSegFlow, a flow matching-based model for 3D curvilinear structure segmentation. The model learns a continuous transformation from a simple source distribution to the target vascular representation, enabling progressive refinement of complex curvilinear geometries with efficient inference. We evaluate our method on Three public challenging datasets covering distinct anatomies and modalities: portal vein, cerebral vessel, and coronary arteries. Using a common architecture and training strategy across all tasks, our method outperforms general-purpose and vessel-specific approaches, with strong preservation of thin branches and vascular continuity. This work not only advances the state-of-the-art in 3D curvilinear segmentation but also opens new avenues for efficient, generalizable, and clinically applicable methods in medical image analysis.


\end{abstract}
\section{Introduction}

Accurate segmentation of curvilinear structures in 3D medical images is essential for computational anatomy and computer-aided diagnosis. Portal, hepatic, coronary, and cerebral vessels exhibit complex branching geometries, caliber variations, and intricate topologies \cite{jin2025aortic}. These characteristics are critical for understanding vascular pathologies (e.g. aneurysms, stenosis, Alzheimer’s disease), surgical planning, and minimally invasive treatments \cite{asrani2019burden,jin2025aortic}. Yet automatic segmentation remains difficult due to several factors \cite{jin2025aortic}:
(i) Severe class imbalance. (vascular networks typically occupy a small fraction of the image volume), (ii) Morphological variability across patients (branching patterns, vessel calibers, tortuosity, and spatial organization), (iii) Imaging complexities: (low contrast between vessels and parenchyma, partial volume effects, blurring boundaries, intensity distortions), and (iv) Manual annotation challenges (time-consuming, inter-annotator variability, inconsistent labels). These challenges hinder the generalization of segmentation algorithms trained on specific vascular anatomies, raising the question of whether a unified approach can achieve robust performance across diverse 3D curvilinear structures.

Over recent years, 3D deep learning has advanced curvilinear segmentation by leveraging full volumetric information. The 3D U-Net extended the U-Net model into three dimensions \cite{cciccek20163d}, becoming a standard for volumetric medical segmentation. Residual connections enabled deeper models \cite{milletari2016v,yu2019liver}, and attention-based methods such as 3D Attention U-Net improved focus on relevant vessel regions while reducing background interference \cite{oktay2018attention}. Transformer-based architectures like UNETR \cite{hatamizadeh2022unetr} and SwinUNETR \cite{hatamizadeh2021swin} further introduced attention to capture long-range dependencies, including some specialized methods in curvilinear structure segmentation like CS\textsuperscript{2}-Net \cite{mou2021cs2} and Lv et al. \cite{lv2023automatic}. More recently, Mamba-based architectures have emerged as an efficient alternative to Transformers by modeling long-range dependencies with linear computational complexity \cite{ma2024u,liu2024mambavesselnet}. However, most methods remain highly specialized, necessitating extensive retraining and limiting generalization.
Foundation models have emerged as promising solutions \cite{wang2025sam}. For example, vesselFM specifically designed for 3D blood vessel segmentation achieved zero-shot generalization across modalities, by training on curated annotations, domain-randomized synthetic data, and flow matching-generated samples \cite{wittmann2025vesselfm}. 
Diffusion-based models demonstrate that iterative refinement improves segmentation by progressively denoising predictions \cite{kazerouni2022diffusion}. However, they require many sampling steps, incurring substantial computational cost, and rely on stochastic processes that introduce variability. Flow matching offers a deterministic alternative, learning a vector field that transports an initial distribution to the target through a continuous trajectory \cite{lipman2022flow,tong2023conditional}. This enables efficient inference with few integration steps. Although 2D flow-based methods have shown promise for progressive refinement and preservation of thin structures \cite{sidelmoctar2026curvsegflow,asadi2026fms}, their extension to 3D vascular segmentation—with increased memory and computational demands—remains largely unexplored and generalization in multiple vascular territories has not been systematically evaluated.

In this work, we introduce 3D-CurvSegFlow, a flow matching-based model for robust 3D curvilinear structure segmentation. Our approach formulates segmentation as a continuous time-dependent process, learning a velocity field that progressively transforms a noisy initialization into the final mask, conditioned on the input image. This iterative refinement enables gradual error correction and maintains structural continuity across vessel networks. The model employs a 3D time-conditioned U-Net with attention-gated skip connections and sinusoidal time embeddings.
Our main contributions are:(1) The first comprehensive evaluation of conditional flow matching for 3D curvilinear structure segmentation across multiple vascular territories, demonstrating robust generalization without task-specific modifications; (2) A time-conditioned 3D U-Net architecture with attention gating and a combined training objective that jointly optimizes trajectory learning and segmentation quality; and (3) Extensive experiments on three challenging vascular datasets, showing that iterative refinement through flow-based modeling consistently improves vessel continuity and reduces topological errors, particularly under challenging imaging conditions.

\section{Methodology}

\subsection{Datasets}

To evaluate generalization across anatomical territories and imaging modalities, we select three publicly available 3D datasets encompassing portal veins, cerebral vasculature, and coronary arteries. These datasets exhibit substantial heterogeneity in acquisition protocols, contrast enhancement, vessel caliber distributions, and annotation quality, providing a rigorous testbed for assessing robustness without task-specific adaptations.

\textbf{3Dircadb (portal vein) \cite{soler20103d}:}
This database comprises 19 contrast-enhanced abdominal CT scans with manual segmentations, from which we target the portal vein. Acquisitions have moderate vessel-to-parenchyma contrast and sparse vascular occupancy, typically less than 2\% of the liver volume. Slice thickness ranges from 1 to 4 mm, with 74 to 260 slices per volume and an in-plane resolution of $512 \times 512$ pixels. We adopt a patient-wise split of 15 volumes for training and 4 for testing.

\textbf{SMILE-UHURA (brain vessels) \cite{chatterjee2024smile}:}
This challenge dataset provides 3D multi-slab time-of-flight MRA data acquired on a 7T Siemens MAGNETOM scanner with an isotropic resolution of 300 $\mu$m. The original collection includes 20 subjects from the StudyForrest project, but only the the training-validation dataset of 14 labelled samples was made available. We randomly partition these into 12 training and 2 test volumes. These volumes feature fine-scale Circle of Willis structures and aneurysms, requiring high sensitivity to thin vessels. 

\textbf{ImageCAS (coronary arteries) \cite{zeng2023imagecas}:}
This large-scale benchmark for coronary artery segmentation comprises 1000 3D CTA volumes, each from a unique patient, acquired with a Siemens 128-slice dual-source scanner. Volumes have dimensions of $512 \times 512 \times (206{-}275)$ voxels. Coronary vessels exhibit severe caliber variations, with the presence of calcified plaques and motion artefacts. We randomly select 800 volumes for training and 200 for test.

\subsection{Proposed Model}
\begin{figure*}
\centering
\includegraphics[width=1.00\linewidth]{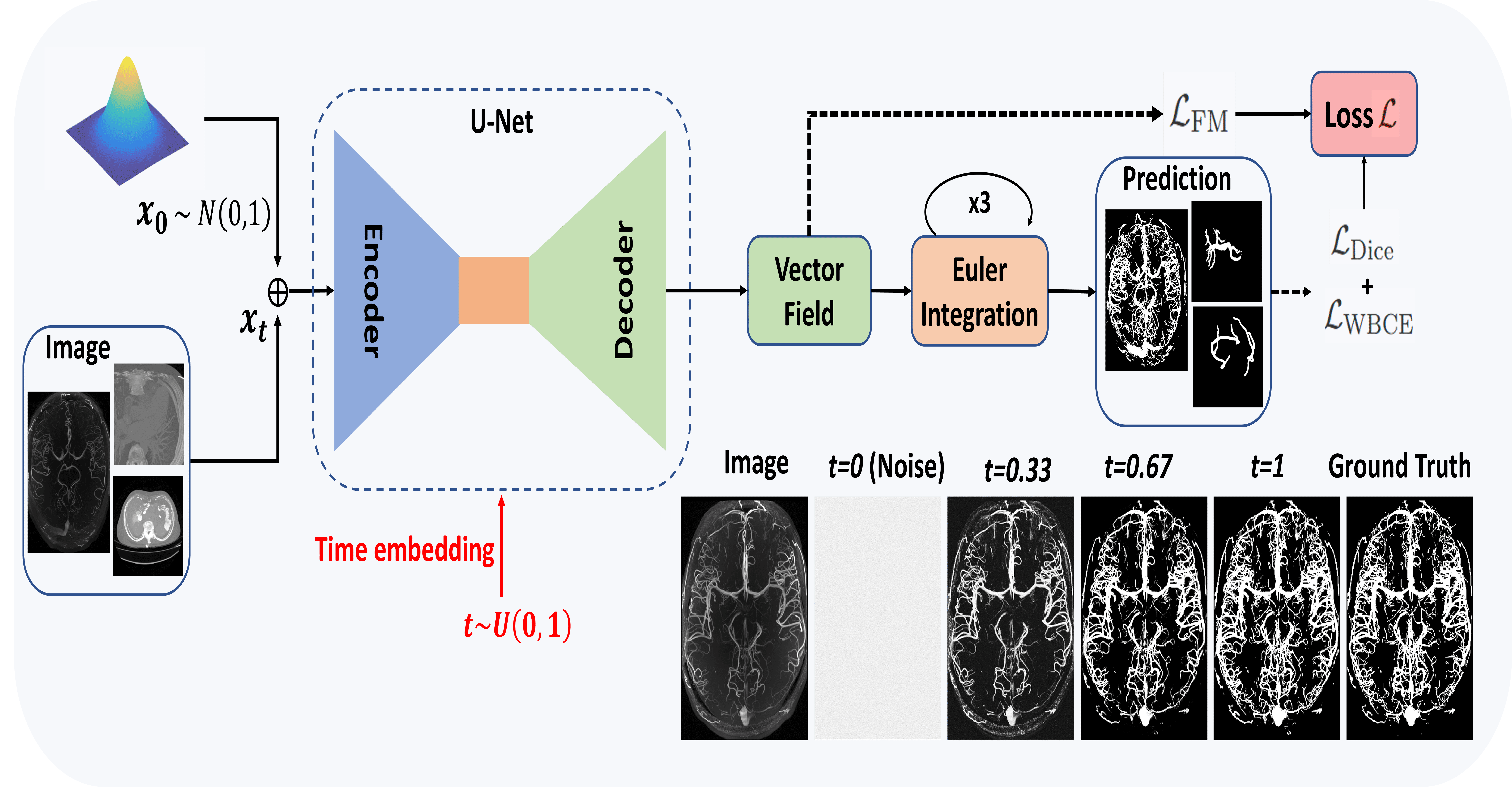}
\caption{\label{fig:3Dflow} \footnotesize An overview of the proposed 3D-CurvSegFlow model}
\end{figure*}

We formulate 3D curvilinear structure segmentation as a continuous dynamical process. Rather than predicting a mask in a single forward pass, the model learns a time-dependent velocity field that progressively transports a noisy initialization toward the ground truth, conditioned on the input image (Figure \ref{fig:3Dflow}).

Let \(I \in \mathbb{R}^{H \times W \times D}\) denote the input volume and \(x_1 \in \{0,1\}^{H \times W \times D}\) the ground-truth binary mask. We sample initial noise \(x_0 \sim \mathcal{N}(0, \mathbf{I})\) independent of \(I\). For continuous time \(t \in [0,1]\), we define a linear interpolation between noise and target:

\begin{equation}
    x_t = (1 - t)x_0 + t x_1
\end{equation}

The corresponding target velocity is \(u_t = x_1 - x_0\). The model learns a vector field \(\nu_\theta(x_t, I, t)\) parameterized by \(\theta\) to predict the instantaneous change of the evolving mask.
We instantiate \(\nu_\theta\) as a 3D time-conditioned U-Net with four encoder levels (channels: 32, 64, 128, 256), a bottleneck (512 channels), and a symmetric decoder. Each convolutional block uses \(3 \times 3 \times 3\) convolutions, Group Normalization, and SiLU activations. Attention gates are applied to all skip connections to focus on vascular regions and suppress background. The total number of parameters is 23.49 million.
Sinusoidal positional embeddings map scalar time \(t\) to a 256-dimensional vector, processed through a two-layer MLP and added element-wise to feature maps at each block. This allows the network to adapt its behaviour across reconstruction stages.
Segmentation is obtained by integrating the learned dynamics from \(t=0\) to \(t=1\) using explicit Euler with \(N=3\) steps (vs. hundreds for diffusion models), followed by a sigmoid \(\hat{x} = \sigma(x_N)\). For each iteration, we sample \(t \sim \mathcal{U}(0,1)\) and \(x_0 \sim \mathcal{N}(0, \mathbf{I})\), compute \(x_t\) and \(u_t\), and update: 

\begin{equation}
    x_{n+1} = x_n + \Delta t \, \nu_\theta(x_n, I, t_n), \quad t_n = n\Delta t, \quad \Delta t = 1/N
\end{equation}
We use a composite loss which includes a flow matching loss:

\begin{equation}
    \mathcal{L}_{\mathrm{FM}} = \frac{1}{2} \| \nu_\theta(x_t, I, t) - u_t \|^2
\end{equation}

To align final reconstruction with ground truth, we apply weighted binary cross-entropy (with class weights \(w_1=1.0, w_0=0.15\)) and soft Dice loss:

\begin{equation}
    \mathcal{L}_{\mathrm{WBCE}} = -\frac{1}{V} \sum [w_1 y \log \hat{x} + w_0 (1-y) \log(1-\hat{x})]
\end{equation}

\begin{equation}
    \mathcal{L}_{\mathrm{Dice}} = 1 - \frac{2\sum y\hat{x} + \epsilon}{\sum y + \sum \hat{x} + \epsilon}
\end{equation}

The total loss is : \(\mathcal{L} = \lambda_{\mathrm{FM}}\mathcal{L}_{\mathrm{FM}} + \lambda_{\mathrm{WBCE}}\mathcal{L}_{\mathrm{WBCE}} + \lambda_{\mathrm{Dice}}\mathcal{L}_{\mathrm{Dice}}\) 

The model is optimized using AdamW with a learning rate of \(5\times10^{-5}\) and a weight decay of \(1\times10^{-5}\). A cosine annealing scheduler gradually reduces the learning rate to \(10^{-6}\). During training, data augmentation is applied through random axis flips, affine transformations (scale variation up to 10\% and rotations up to \(10^\circ\)), gamma intensity adjustments, and Gaussian noise injection to improve generalization. Training is performed for up to 700 epochs for 3Dircadb and SMILE-UHURA and up to 200 for ImageCAS, with early stopping based on an exponential moving average (EMA) of the validation loss (\(\alpha=0.1\)).

\subsection{Metrics}
Quantitative evaluation employs six metrics to assess both volumetric overlap and topological fidelity. Dice and Intersection over Union (IoU) serve as primary measures of spatial agreement, with Dice being particularly well-suited for imbalanced vascular segmentation. Precision and recall evaluated the trade-off between over-segmentation and under-segmentation, which is critical for minimizing false positives while maintaining sensitivity. To assess topological preservation, we use centerline Dice (clDice), which quantifies overlap of predicted and ground truth centerlines and is sensitive to broken branches, and 95th percentile Hausdorff Distance (HD95), which provides a robust measure of boundary accuracy that is insensitive to outliers. These topological metrics are essential for curvilinear structures where preserving continuity and branching patterns is clinically more relevant than pixel-wise overlap alone.

\section{Results and Discussion}
We compare the segmentation results of 3D-CurvSegFlow with recent state-of-the art models. This include $D^{2}$-RD-UNet (2025) for hepatic vessel segmentation \cite{cavicchioli2025d2}; FFCM-MRF (2024) and vsesselFM (2025) for cerebrovascular segmentation \cite{cui2024ffcm,wittmann2025vesselfm}; and MSFP-Net (2026) for coronary artery segmentation \cite{wang2026msfp} (Table \ref{tab:combined_results}). Table \ref{tab:combined_results} also includes the performance of : (i) models from the benchmarks associated with the aforementioned state-of-the-art methods, and (ii) general-purpose models (nnUNet, 3D-UNet and CS2-Net) trained and tested on the same images as 3D-CurvSegFlow. 

Importantly, CurvSegFlow employs a single architecture and identical training strategy to segment various types of curvilinear structures across all datasets without anatomy-specific modifications. This allows the evaluation to focus on the generalizability of the proposed flow matching model.
3D-CurvSegFlow consistently achieves the best Dice, clDice and recall scores across all datasets. It also achives the best IoU, precision and HD95 scores for two out of three datasets. This consistency across portal, cerebral, and coronary vessels is particularly noteworthy because these datasets differ substantially in imaging modality, anatomical complexity, vessel diameter distribution, and annotation characteristics. The results therefore suggest that the proposed formulation captures general properties of curvilinear structures rather than features specific to a particular vascular territory.

\begin{table}[ht]
\centering
\scriptsize
\caption{Quantitative evaluation of vessel segmentation performance across three datasets. We compare 3D-CurvSegFlow to various state-of-the-art models : $^{*}$ Results from \cite{cavicchioli2025d2}; $^{\dagger}$ Results from \cite{wittmann2025vesselfm}: few-shot task; $^{\ddagger}$ Results from \cite{cui2024ffcm}; and $^{\bullet}$ Results from \cite{wang2026msfp}: 5-fold cross-validation performed. Results are reported as mean across up to six metrics, with best values highlighted in bold.}
\begin{tabular}{l|l|ccccccc}
\hline
\textbf{Dataset} & \textbf{Model} & \textbf{Dice$\uparrow$} & \textbf{IoU$\uparrow$} & \textbf{Precision$\uparrow$} & \textbf{Recall$\uparrow$} & \textbf{clDice$\uparrow$} & \textbf{HD95$\downarrow$} \\
\hline
\multirow{8}{*}{\textbf{3Dircadb}} 
& 3D U-Net \cite{cciccek20163d} & 0.36 & 0.224 & 0.29 & 0.49 & 0.36 & 71.62 \\
& CS\textsuperscript{2}-Net \cite{mou2021cs2} & 0.40 & 0.259 & 0.32 & 0.59 & 0.41 & 40.46 \\
& nnU-Net \cite{isensee2021nnu} & 0.70 & 0.544 & 0.74 & 0.67 & 0.64 & 20.85 \\
& VNet$^{*}$ \cite{milletari2016v} & 0.61 & 0.439 & 0.66 & 0.60 & 0.60 & 23.59 \\
& Att-UNet$^{*}$ \cite{oktay2018attention} & 0.61 & 0.439 & 0.64 & 0.62 & 0.60 & 24.43 \\
& SwinUNETR$^{*}$ \cite{hatamizadeh2021swin} & 0.61 & 0.439 & 0.68 & 0.60 & 0.62 & 24.82 \\
& $D^{2}$-RD-UNet$^{*}$ \cite{cavicchioli2025d2} & 0.66 & 0.493 & 0.67 & 0.68 & 0.68 & 23.31 \\
& \textbf{Ours} & \textbf{0.74} & \textbf{0.587} & \textbf{0.78} & \textbf{0.71} & \textbf{0.70} & \textbf{18.54} \\
\hline
\multirow{8}{*}{\textbf{SMILE-UHURA}} 
& 3D U-Net \cite{cciccek20163d} & 0.5121 & 0.4330 & 0.5243 & 0.5116 & 0.4815 & 5.00 \\
& CS\textsuperscript{2}-Net \cite{mou2021cs2} & 0.5118 & 0.4158 & 0.5026 & 0.5318 & 0.4804 & 5.70 \\
& nnU-Net \cite{isensee2021nnu} & 0.7981 & 0.6702 & 0.9238 & 0.7168 & 0.8294 & \textbf{3.38} \\
& GCS$^{\ddagger}$ \cite{chen2022generative} & 0.7082 & 0.5482 & \textbf{0.9346} & 0.5789 & -- & 56.47 \\
& GMM-MRF$^{\ddagger}$ \cite{zhang2019device} & 0.7644 & 0.6188 & 0.8166 & 0.7370 & -- & 54.19 \\
& FFCM-MRF$^{\ddagger}$ \cite{cui2024ffcm} & 0.7706 & 0.6267 & 0.8416 & 0.7273 & -- & 57.17 \\
& SAM-Med3D$^{\dagger}$ \cite{wang2025sam} & 0.4659 & 0.3037 & -- & -- & 0.4463 & -- \\
& vesselFM$^{\dagger}$ \cite{wittmann2025vesselfm} & 0.5815 & 0.4100 & -- & -- & 0.4272 & -- \\
& \textbf{Ours} & \textbf{0.8032} & \textbf{0.6763} & 0.8780 & \textbf{0.7734} & \textbf{0.8418} & 3.87 \\
\hline
\multirow{10}{*}{\textbf{ImageCAS}} 
& 3D U-Net \cite{cciccek20163d} & 0.7231 & 0.5718 & 0.6464 & 0.8250 & 0.7549 & 75.16 \\
& nnU-Net \cite{isensee2021nnu} & 0.8059 & 0.6783 & 0.7866 & 0.8325 & 0.8721 & 19.58 \\
& VNet$^{\bullet}$ \cite{milletari2016v} & 0.7305 & 0.575 & 0.7275 & 0.7669 & -- & 23.01 \\
& CAS-Net$^{\bullet}$ \cite{dong2023novel} & 0.7077 & 0.548 & 0.7419 & 0.7128 & -- & 31.87 \\
& UNETR$^{\bullet}$ \cite{hatamizadeh2022unetr} & 0.7375 & 0.584 & 0.7121 & 0.7744 & -- & 21.96 \\
& Swin-UNETR$^{\bullet}$ \cite{hatamizadeh2021swin} & 0.7454 & 0.594 & 0.7189 & 0.7817 & -- & 20.57 \\
& SegFormer$^{\bullet}$ \cite{xie2021segformer} & 0.7315 & 0.577 & 0.7234 & 0.7532 & -- & 18.76 \\
& MSFP-Net$^{\bullet}$ \cite{wang2026msfp} & 0.7923 & 0.656 & 0.7974 & 0.7924 & -- & \textbf{11.25} \\
& \textbf{Ours} & \textbf{0.824} & \textbf{0.701} & \textbf{0.8020} & \textbf{0.850} & \textbf{0.8793} & 22.62 \\
\hline
\end{tabular}
\label{tab:combined_results}
\end{table}

The improvements are especially pronounced in terms of the recall-precision balance and clDice (altough no comparison done was performed on imageCAS for the latter). This observation is clinically important because vascular segmentation errors rarely originate from inaccurate delineation of large vessels. Instead, they mainly arise from missed peripheral branches and broken vessel connectivity. The higher recall indicates that the proposed model detects a larger proportion of small vessels, while the hight precision reflects that this improved detection does not come at the cost of increased false positives. Competitive clDice scores further demonstrate that these additional detections preserve the continuity of the vascular tree, rather than introducing fragmented predictions. These results are consistent with the design of the proposed model, where segmentation is formulated as a continuous transport process instead of a single-step voxel classification problem. During integration, each refinement step takes advantage of both the image appearance and the evolving segmentation state, enabling the network to progressively correct incomplete vessel trajectories and recover thin ambiguous branches.

\begin{figure*}
\centering
\includegraphics[width=0.99\linewidth]{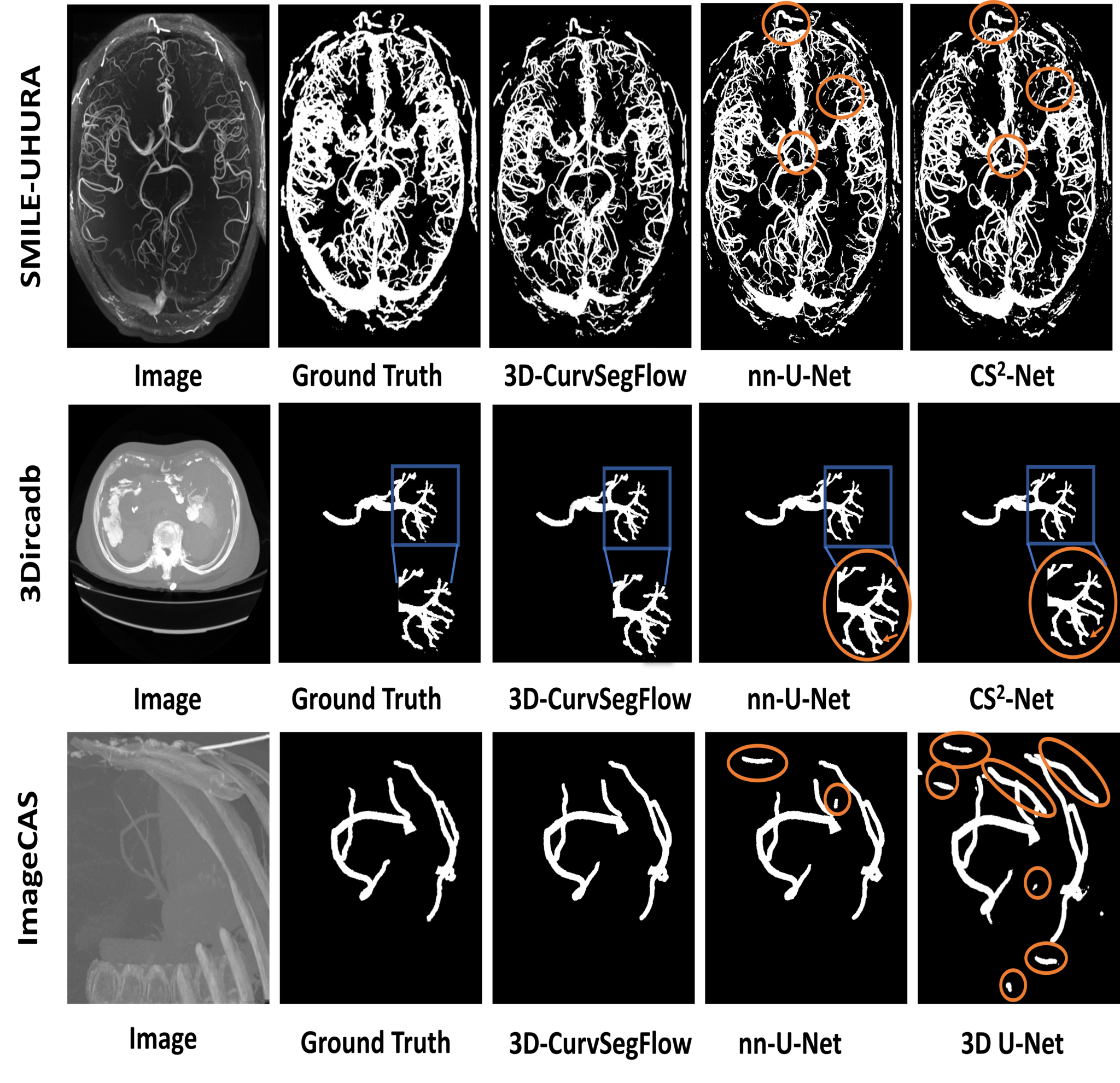}
\caption{\label{fig:3Dflow_pred} \footnotesize Qualitative comparison of 3D-CurvSegFlow, nnU-Net, CS\textsuperscript{2}-Net and 3D U-Net on SMILE-UHURA, 3Dircadb, and ImageCAS. The orange circles highlight regions where predictions are incorrect. }
\end{figure*}

The qualitative examples shown in Figure~\ref{fig:3Dflow_pred} further support the quantitative evaluation. The proposed method generates more continuous vascular trees with fewer disconnected distal branches. These differences are particularly visible in regions affected by weak contrast or imaging artifacts, where competing methods frequently interrupt vessel trajectories or produce false positives, while 3D-CurvSegFlow maintains anatomical continuity and accurate segmentation. A similar trend is observed for cerebral vessels, where numerous fine branches recovered by the proposed model remain absent in competing predictions. Although these local improvements may represent only a limited number of voxels, they have a disproportionately large impact on vascular topology and may be clinically important for downstream applications such as surgical planning, vessel quantification, and computational vascular analysis. Notably, the annotations in the SMILE-UHURA dataset appear wider than the vessels visible in the images. Interestingly, 3D-CurvSegFlow’s predictions more accurately capture the true vessel width, outperforming the manual annotations and illustrating its enhanced capabilities. 
The proposed method consistently outperforms existing approaches across datasets, demonstrating robust generalization under varying imaging conditions. Improvements are observed in challenging scenarios characterized by severe class imbalance, thin vessels, and imaging artifacts. This suggests that modeling segmentation as a continuous flow better preserves vessel continuity and topology than conventional one-step prediction. 

However, some limitations should be acknowledged. First, the evaluation is performed on three vascular datasets acquired using contrast-enhanced CT or high-resolution TOF-MRA. The model’s behavior on low- or non-contrast modalities remains untested. Second, training relies on fixed-size volumetric patches, which may restrict access to global anatomical context for very large vascular trees. Third, although the proposed method substantially improves topological preservation, occasional distal over-segmentation still affects boundary-based metrics such as HD95 on coronary CTA. 
Finally,  we demonstrate computational efficiency by using only three Euler integration steps, achieving 22.3 s per volume on SMILE-UHURA during inference with a general Dice score of 0.8032. Testing 5 and 10 inference steps not only increases inference time but also degrades prediction accuracy: 5 steps: 36.4 s, Dice 0.773; 10 steps: 71.9 s, Dice 0.753. This reveals that 3 inference steps are sufficient and orders of magnitude more efficient than the hundreds of steps typically required by diffusion models. However, additional investigation of alternative ODE solvers and adaptive integration strategies may further improve the accuracy-inference speed trade-off.

\section{Conclusion}
This paper introduces 3D-CurvSegFlow, a flow matching-based framework for robust 3D segmentation of curvilinear anatomical structures. By formulating segmentation as a continuous dynamical process, the proposed method learns a time-conditioned vector field that progressively refines the segmentation while preserving vascular topology using only a few integration steps. Experiments on three public datasets covering different anatomies and imaging modalities demonstrate consistent improvements over other methods, including vessel-specific methods (e.g. $D^{2}$-RD-UNet, CS\textsuperscript{2}-Net, or MSFP-Net), and recent foundation models (e.g. VesselFM). These results highlight the robustness and generalization capability of the proposed model and suggest that conditional flow matching provides an effective and computationally efficient paradigm for 3D curvilinear structure segmentation. In future work, we will investigate its extension to additional imaging modalities, larger datasets, and adaptive integration strategies to further improve scalability and boundary accuracy in complex clinical scenarios.

\clearpage
\bibliographystyle{splncs04}
\bibliography{ref}

@article{asrani2019burden,
  title={Burden of liver diseases in the world},
  author={Asrani, Sumeet K and Devarbhavi, Harshad and Eaton, John and Kamath, Patrick S},
  journal={Journal of hepatology},
  volume={70},
  number={1},
  pages={151--171},
  year={2019},
  publisher={Elsevier}
}

@article{jin2025aortic,
  title={Aortic vessel tree segmentation for cardiovascular diseases treatment: Status quo},
  author={Jin, Yuan and Pepe, Antonio and Li, Jianning and Gsaxner, Christina and Chen, Yuxuan and Puladi, Behrus and Zhao, Fen-hua and Pomykala, Kelsey and Kleesiek, Jens and Frangi, Alejandro F and others},
  journal={ACM Computing Surveys},
  volume={57},
  number={9},
  pages={1--35},
  year={2025},
  publisher={ACM New York, NY}
}

@inproceedings{cciccek20163d,
  title={3D U-Net: learning dense volumetric segmentation from sparse annotation},
  author={{\c{C}}i{\c{c}}ek, {\"O}zg{\"u}n and Abdulkadir, Ahmed and Lienkamp, Soeren S and Brox, Thomas and Ronneberger, Olaf},
  booktitle={International conference on medical image computing and computer-assisted intervention},
  pages={424--432},
  year={2016},
  organization={Springer}
}

@article{ma2024u,
  title={U-mamba: Enhancing long-range dependency for biomedical image segmentation},
  author={Ma, Jun and Li, Feifei and Wang, Bo},
  journal={arXiv preprint arXiv:2401.04722},
  year={2024}
}

@article{dong2023novel,
  title={A novel multi-attention, multi-scale 3D deep network for coronary artery segmentation},
  author={Dong, Caixia and Xu, Songhua and Dai, Duwei and Zhang, Yizhi and Zhang, Chunyan and Li, Zongfang},
  journal={Medical Image Analysis},
  volume={85},
  pages={102745},
  year={2023},
  publisher={Elsevier}
}

@inproceedings{hatamizadeh2022unetr,
  title={Unetr: Transformers for 3d medical image segmentation},
  author={Hatamizadeh, Ali and Tang, Yucheng and Nath, Vishwesh and Yang, Dong and Myronenko, Andriy and Landman, Bennett and Roth, Holger R and Xu, Daguang},
  booktitle={Proceedings of the IEEE/CVF winter conference on applications of computer vision},
  pages={574--584},
  year={2022}
}

@article{oktay2018attention,
  title={Attention u-net: Learning where to look for the pancreas},
  author={Oktay, Ozan and Schlemper, Jo and Folgoc, Loic Le and Lee, Matthew and Heinrich, Mattias and Misawa, Kazunari and Mori, Kensaku and McDonagh, Steven and Hammerla, Nils Y and Kainz, Bernhard and others},
  journal={arXiv preprint arXiv:1804.03999},
  year={2018}
}

@inproceedings{yu2019liver,
  title={Liver vessels segmentation based on 3d residual U-NET},
  author={Yu, Wei and Fang, Bin and Liu, Yongqing and Gao, Mingqi and Zheng, Shenhai and Wang, Yi},
  booktitle={2019 IEEE international conference on image processing (ICIP)},
  pages={250--254},
  year={2019},
  organization={IEEE}
}

@article{mou2021cs2,
  title={CS2-Net: Deep learning segmentation of curvilinear structures in medical imaging},
  author={Mou, Lei and Zhao, Yitian and Fu, Huazhu and Liu, Yonghuai and Cheng, Jun and Zheng, Yalin and Su, Pan and Yang, Jianlong and Chen, Li and Frangi, Alejandro F and others},
  journal={Medical image analysis},
  volume={67},
  pages={101874},
  year={2021},
  publisher={Elsevier}
}

@article{kazerouni2022diffusion,
  title={Diffusion models for medical image analysis: A comprehensive survey},
  author={Kazerouni, Amirhossein and Aghdam, Ehsan Khodapanah and Heidari, Moein and Azad, Reza and Fayyaz, Mohsen and Hacihaliloglu, Ilker and Merhof, Dorit},
  journal={arXiv preprint arXiv:2211.07804},
  year={2022}
}

@article{lipman2022flow,
  title={Flow matching for generative modeling},
  author={Lipman, Yaron and Chen, Ricky TQ and Ben-Hamu, Heli and Nickel, Maximilian and Le, Matt},
  journal={arXiv preprint arXiv:2210.02747},
  year={2022}
}

@article{tong2023conditional,
  title={Conditional flow matching: Simulation-free dynamic optimal transport},
  author={Tong, Alexander and Malkin, Nikolay and Huguet, Guillaume and Zhang, Yanlei and Rector-Brooks, Jarrid and Fatras, Kilian and Wolf, Guy and Bengio, Yoshua},
  journal={arXiv preprint arXiv:2302.00482},
  volume={2},
  number={3},
  year={2023}
}

@inproceedings{wittmann2025vesselfm,
  title={vesselfm: A foundation model for universal 3d blood vessel segmentation},
  author={Wittmann, Bastian and Wattenberg, Yannick and Amiranashvili, Tamaz and Shit, Suprosanna and Menze, Bjoern},
  booktitle={Proceedings of the IEEE/CVF Conference on Computer Vision and Pattern Recognition},
  pages={20874--20884},
  year={2025}
}

@article{cavicchioli2025d2,
  title={D2-RD-UNet: A dual-stage dual-class framework with connectivity correction for hepatic vessels segmentation},
  author={Cavicchioli, Matteo and Moglia, Andrea and Garret, Guillaume and Puglia, Martina and Vacavant, Antoine and Pugliese, Giacomo and Cerveri, Pietro},
  journal={Computers in Biology and Medicine},
  volume={195},
  pages={110530},
  year={2025},
  publisher={Elsevier}
}

@article{zeng2023imagecas,
  title={ImageCAS: A large-scale dataset and benchmark for coronary artery segmentation based on computed tomography angiography images},
  author={Zeng, An and Wu, Chunbiao and Lin, Guisen and Xie, Wen and Hong, Jin and Huang, Meiping and Zhuang, Jian and Bi, Shanshan and Pan, Dan and Ullah, Najeeb and others},
  journal={Computerized Medical Imaging and Graphics},
  volume={109},
  pages={102287},
  year={2023},
  publisher={Elsevier}
}

@article{chatterjee2024smile,
  title={SMILE-UHURA Challenge--Small Vessel Segmentation at Mesoscopic Scale from Ultra-High Resolution 7T Magnetic Resonance Angiograms},
  author={Chatterjee, Soumick and Mattern, Hendrik and D{\"o}rner, Marc and Sciarra, Alessandro and Dubost, Florian and Schnurre, Hannes and Khatun, Rupali and Yu, Chun-Chih and Hsieh, Tsung-Lin and Tsai, Yi-Shan and others},
  journal={arXiv preprint arXiv:2411.09593},
  year={2024}
}

@article{soler20103d,
  title={3d image reconstruction for comparison of algorithm database},
  author={Soler, Luc and Hostettler, Alexandre and Agnus, Vincent and Charnoz, Arnaud and Fasquel, Jean-Baptiste and Moreau, Johan and Osswald, Anne-Blandine and Bouhadjar, Mourad and Marescaux, Jacques},
  journal={URL: https://www. ircad. fr/research/data-sets/liver-segmentation-3d-ircadb-01},
  volume={13},
  pages={4},
  year={2010}
}

@article{cui2024ffcm,
  title={FFCM-MRF: An accurate and generalizable cerebrovascular segmentation pipeline for humans and rhesus monkeys based on TOF-MRA},
  author={Cui, Yue and Huang, Haibin and Liu, Jialu and Zhao, Mingyang and Li, Chengyi and Han, Xinyong and Luo, Na and Gao, Jinquan and Yan, Dong-Ming and Zhang, Chen and others},
  journal={Computers in Biology and Medicine},
  volume={170},
  pages={107996},
  year={2024},
  publisher={Elsevier}
}

@inproceedings{zhang2019device,
  title={A device-independent novel statistical modeling for cerebral TOF-MRA data segmentation},
  author={Zhang, Baochang and Wu, Zonghan and Liu, Shuting and Zhou, Shoujun and Li, Na and Zhao, Gang},
  booktitle={Workshop on Clinical Image-Based Procedures},
  pages={172--181},
  year={2019},
  organization={Springer}
}

@article{chen2022generative,
  title={Generative consistency for semi-supervised cerebrovascular segmentation from TOF-MRA},
  author={Chen, Cheng and Zhou, Kangneng and Wang, Zhiliang and Xiao, Ruoxiu},
  journal={IEEE Transactions on Medical Imaging},
  volume={42},
  number={2},
  pages={346--353},
  year={2022},
  publisher={IEEE}
}

@article{lv2023automatic,
  title={Automatic segmentation of mandibular canal using transformer based neural networks},
  author={Lv, Jinxuan and Zhang, Lang and Xu, Jiajie and Li, Wang and Li, Gen and Zhou, Hengyu},
  journal={Frontiers in Bioengineering and Biotechnology},
  volume={11},
  pages={1302524},
  year={2023},
  publisher={Frontiers Media SA}
}

@article{liu2024mambavesselnet,
  title={Mambavesselnet: a novel approach to blood vessel segmentation based on state-space models},
  author={Liu, Tianyong and Zhang, Zhiqing and Fan, Guojia and Li, Bin and Zhou, Shoujun and Xu, Chengwu and Zhao, Gang and Yang, Fuxia},
  journal={IEEE Journal of Biomedical and Health Informatics},
  volume={29},
  number={3},
  pages={2034--2047},
  year={2024},
  publisher={IEEE}
}

@article{isensee2021nnu,
  title={nnU-Net: a self-configuring method for deep learning-based biomedical image segmentation},
  author={Isensee, Fabian and Jaeger, Paul F and Kohl, Simon AA and Petersen, Jens and Maier-Hein, Klaus H},
  journal={Nature methods},
  volume={18},
  number={2},
  pages={203--211},
  year={2021},
  publisher={Nature Publishing Group US New York}
}

@inproceedings{hatamizadeh2021swin,
  title={Swin unetr: Swin transformers for semantic segmentation of brain tumors in mri images},
  author={Hatamizadeh, Ali and Nath, Vishwesh and Tang, Yucheng and Yang, Dong and Roth, Holger R and Xu, Daguang},
  booktitle={International MICCAI brainlesion workshop},
  pages={272--284},
  year={2021},
  organization={Springer}
}

@article{wang2026msfp,
  title={MSFP-Net: a multi-scale and multi-frequency enhanced 3D network with learnable priors for coronary artery segmentation in CCTA images},
  author={Wang, Ankang and Wang, Lu and Xu, Lisheng and Liu, Jinshuai and Sun, Yu and Zhang, Libo},
  journal={Biomedical Signal Processing and Control},
  volume={115},
  pages={109366},
  year={2026},
  publisher={Elsevier}
}

@article{xie2021segformer,
  title={SegFormer: Simple and efficient design for semantic segmentation with transformers},
  author={Xie, Enze and Wang, Wenhai and Yu, Zhiding and Anandkumar, Anima and Alvarez, Jose M and Luo, Ping},
  journal={Advances in neural information processing systems},
  volume={34},
  pages={12077--12090},
  year={2021}
}

@article{wang2025sam,
  title={SAM-Med3D: a vision foundation model for general-purpose segmentation on volumetric medical images},
  author={Wang, Haoyu and Guo, Sizheng and Ye, Jin and Deng, Zhongying and Cheng, Junlong and Li, Tianbin and Chen, Jianpin and Su, Yanzhou and Huang, Ziyan and Shen, Yiqing and others},
  journal={IEEE Transactions on Neural Networks and Learning Systems},
  year={2025},
  publisher={IEEE}
}

@article{asadi2026fms,
  title={FMS$^{2}$: Unified Flow Matching for Segmentation and Synthesis of Thin Structures},
  author={Asadi, Babak and Wu, Peiyang and Golparvar-Fard, Mani and Shah, Viraj and Hajj, Ramez},
  journal={arXiv preprint arXiv:2603.13659},
  year={2026}
}

@article{sidelmoctar2026curvsegflow,
  title={CurvSegFlow: Time-Conditioned Flow Matching for Robust Segmentation of Curvilinear Structures in Noisy Biomedical Images},
  author={
    {Sid'El Moctar}, Sidi Mohamed and
    {Ait Laydi}, Achraf and
    Beber, Alexandre and
    Braun, Marcus and
    Lansky, Zdenek and
    {El Mourabit}, Yousef and
    Bouvrais, H{\'e}l{\`e}ne
  },
  journal={arXiv preprint arXiv:2606.21608},
  year={2026}
}

@inproceedings{milletari2016v,
  title={V-net: Fully convolutional neural networks for volumetric medical image segmentation},
  author={Milletari, Fausto and Navab, Nassir and Ahmadi, Seyed-Ahmad},
  booktitle={2016 fourth international conference on 3D vision (3DV)},
  pages={565--571},
  year={2016},
  organization={Ieee}
}
\end{document}